\documentclass[letterpaper]{article} 
\usepackage{aaai23}  
\usepackage{times}  
\usepackage{helvet}  
\usepackage{courier}  
\usepackage[hyphens]{url}  
\usepackage{graphicx} 
\usepackage{natbib}  
\usepackage{caption} 
\usepackage{amsmath}
\usepackage{amsmath}
\usepackage{graphicx}
\usepackage{amsmath}
\usepackage{amssymb}
\usepackage{booktabs}
\usepackage{multirow}
\usepackage{algorithm}
\usepackage{algorithmic}
\usepackage{xcolor}
\usepackage{graphicx}
\usepackage{algorithm}
\usepackage{algorithmic}
\usepackage{tikz}
\usepackage{comment}
\usepackage{amsmath,amssymb}
\usepackage{multirow}

\usepackage{newfloat}
\usepackage{listings}
\DeclareCaptionStyle{ruled}{labelfont=normalfont,labelsep=colon,strut=off} 
\floatstyle{ruled}
\newfloat{listing}{tb}{lst}{}
\floatname{listing}{Listing}
\title{Recurrent Structure Attention Guidance for Depth Super-Resolution}
\author{
    Jiayi Yuan\equalcontrib, Haobo Jiang\equalcontrib, Xiang Li, Jianjun Qian\footnotemark[2], Jun Li\thanks{corresponding authors}, Jian Yang
}
\affiliations{
    PCA Lab, Key Lab of Intelligent Perception and Systems for High-Dimensional Information of Ministry of Education\\
    Jiangsu Key Lab of Image and Video Understanding for Social Security\\
    School of Computer Science and Engineering, Nanjing University of Science and Technology, Nanjing, China\\
     
\{jiayiyuan, jiang.hao.bo, xiang.li.implus, csjqian, junli, csjyang\}@njust.edu.cn
}

\usepackage{bibentry}

\begin{document}

\maketitle

\begin{abstract}
Image guidance is an effective strategy for depth super-resolution. Generally, most existing methods employ hand-crafted operators to decompose the high-frequency (HF) and low-frequency (LF) ingredients from low-resolution depth maps and guide the HF ingredients by directly concatenating them with image features. However, the hand-designed operators usually cause inferior HF maps (e.g., distorted or structurally missing) due to the diverse appearance of complex depth maps. Moreover, the direct concatenation often results in weak guidance because not all image features have a positive effect on the HF maps. In this paper, we develop a recurrent structure attention guided (RSAG) framework, consisting of two important parts. First, we introduce a deep contrastive network with multi-scale filters for adaptive frequency-domain separation, which adopts contrastive networks from large filters to small ones to calculate the pixel contrasts for adaptive high-quality HF predictions. Second, instead of the coarse concatenation guidance, we propose a recurrent structure attention block, which iteratively utilizes the latest depth estimation and the image features to jointly select clear patterns and boundaries, aiming at providing refined guidance for accurate depth recovery. In addition, we fuse the features of HF maps to enhance the edge structures in the decomposed LF maps. Extensive experiments show that our approach obtains superior performance compared with state-of-the-art depth super-resolution methods.
\end{abstract}

\section{Introduction}

\begin{figure}[t]
  \centering
     \includegraphics[width=1.0\linewidth]{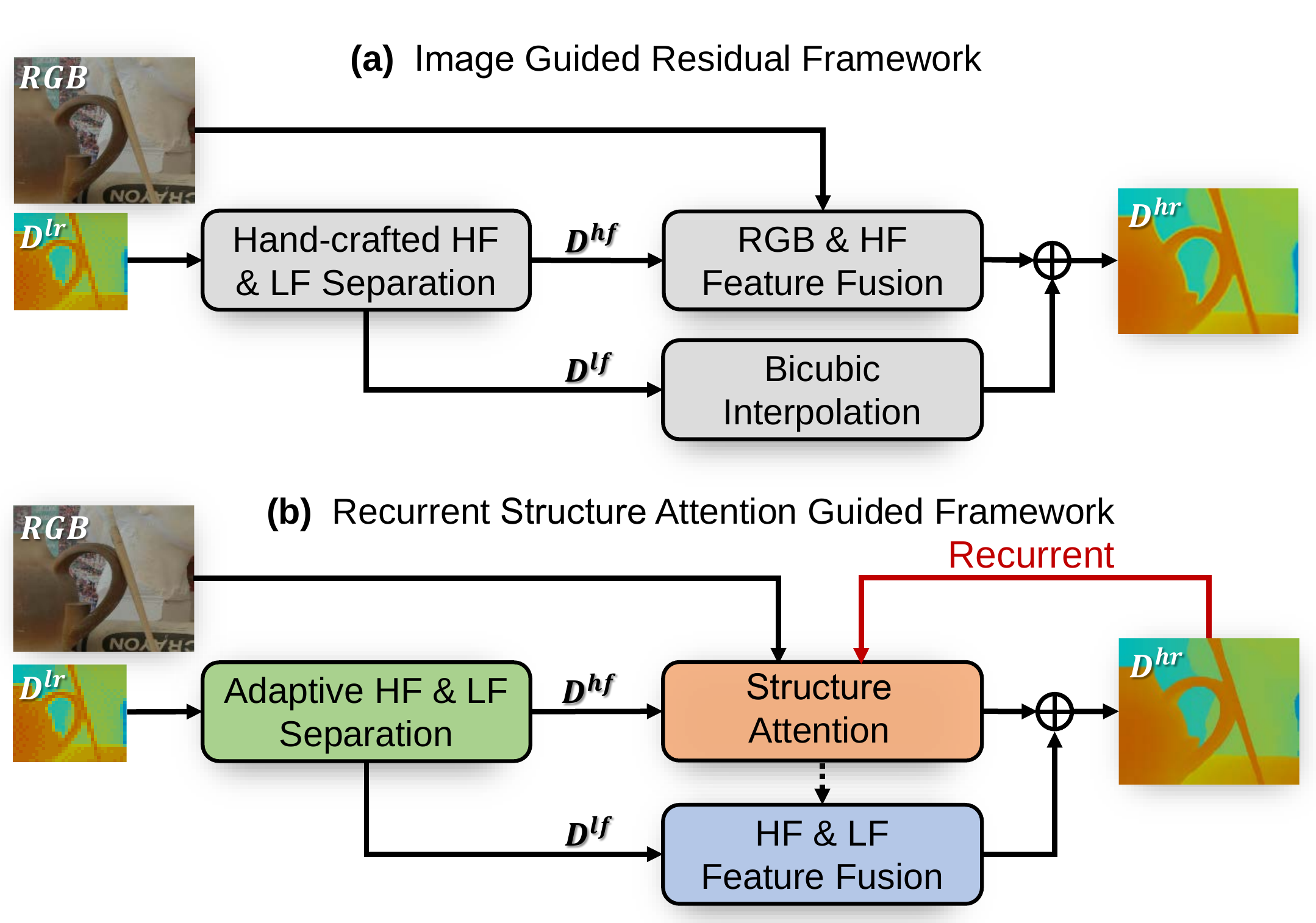}
   \caption{Image guided DSR framework. (a) Popular image guided residual DSR framework; (b) Our recurrent structure attention guided DSR framework.}
   \label{fig:dsrframework}
\end{figure}

Depth super-resolution (DSR) is a fundamental low-level vision topic in computer vision as it plays an important role in a variety of applications, such as 3D reconstruction \cite{hou20193d}, autonomous driving \cite{caesar2020nuscenes}, and virtual reality \cite{meuleman2020single}. Generally, DSR is to recover a high-resolution depth map precisely from a given low-resolution depth map. Recently, an image guidance DSR framework becomes more and more popular since it has demonstrated remarkable progress by borrowing the structures and boundaries in high-resolution image to improve the depth map \cite{2016Multi}. As shown in Fig.~\ref{fig:dsrframework} (a), most efforts
\cite{2016Multi,2019Hierarchical,zuo2019multi,li2019joint} usually 1) use hand-crafted operators (e.g., hand-designed filters) to perform an early spectral decomposition of the low-resolution depth map (i.e., high-frequency (HF) and low-frequency (LF)), 2) coarsely implement the image guidance by directly concatenating the image features into the HF maps, and 3) run a simple up-sampling like bicubic interpolation on the decomposed low-resolution LF map to a high-resolution one. However, this framework still suffers from three challenging problems as follows.

\begin{figure*}[t]
  \centering
    \includegraphics[width=0.80\linewidth]{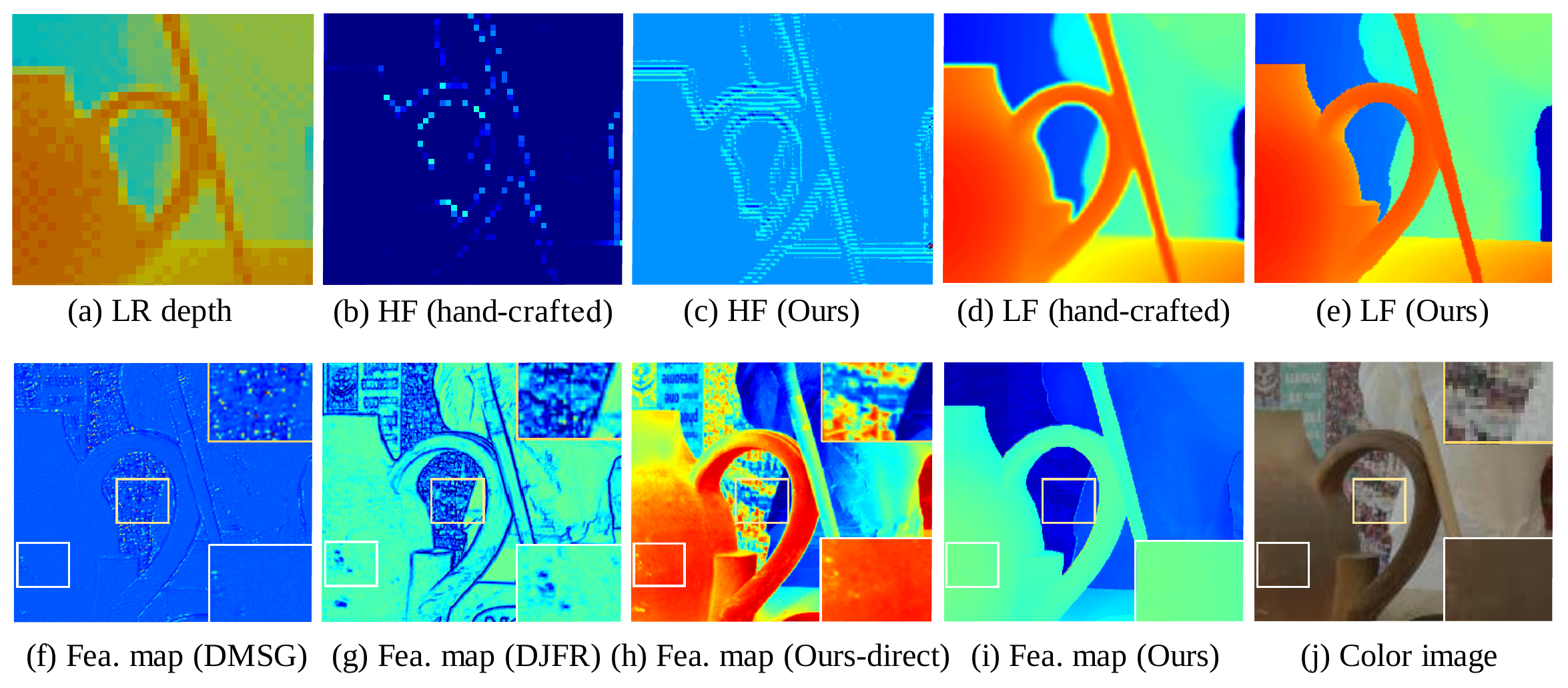}
    \caption{Visualizations of the decomposed HF\&LF  and guidance feature maps. (b) and (d) show a weak frequency-domain separation using the hand-designed operators\cite{2016Multi}. (f-h) show image guidance features with redundant textures and noise in DMSG \cite{2016Multi}, DJFR \cite{li2019joint} and our network using direct concatenation. Compared with them, our method produces better HF structure in (c), sharper LF boundaries in (e), and clearer guidance structure in (i). LR depth map and color image are plotted in (a) and (g).}
    \label{fig:problems}
\end{figure*} 

Firstly, the hand-designed operators often cause a weak spectral decomposition as they are difficult to handle the diverse structures in the complex depth map, resulting in lost object structures in the HF map of Fig.~\ref{fig:problems} (b). Secondly, the direct feature concatenation results in weak image guidance since the complex textures of the image usually produce inferior features based on our observation. For example, Fig.~\ref{fig:problems} (f-h) show clear features of the ceramic bottle (white box) and poor features of the poster (yellow box), corresponding to the complex and simple textures of the image in Fig.~\ref{fig:problems} (j), respectively. Thirdly, the bicubic interpolation also results in blurred edges in the LF map of Fig.~\ref{fig:problems} (d), because it is unsuitable for up-sampling of all kinds of structures.

To address these problems, we develop a novel recurrent structure attention guided (RSAG) framework for high-quality DSR in Fig.~\ref{fig:dsrframework} (b) through three aspects. First of all, we introduce a deep contrastive network with multi-scale filters (DCN) to effectively decompose the HF and LF components of the input depth, instead of the hand-designed operators. DCN is to subtly stack simple contrastive networks \cite{xu2020learning} three times from large filters to small ones for a coarse-to-fine HF prediction with contextual structures, and to calculate the LF component by subtracting the HF prediction from the input depth map. To better guide the depth features, in addition, we propose a recurrent structure attention (SA) block to select the useful image features, instead of the direct concatenation guidance. The key step of SA is to add absolute values of contrastive attention features of the image and the latest depth prediction, and then calculate an attention map by employing channel and spatial attention operators. Finally, we present an HF$\&$LF feature fusion (HLF) block to improve the blurred edges in the LF component by concatenating the HF feature produced by our SA block, as its contextual structure can enhance the edges. Overall, our RSAG framework has a significant improvement on the hand-crafted spectral decomposition and image guidance. In summary, our contributions are as follows:

\begin{itemize}
\item We introduce a deep contrastive network with multi-scale filters (DCN) for the robust HF and LF reconstruction, where the HF structure is implemented by stacking the pixel-wise contrast from large to small kernels.
\item We propose a novel recurrent structure attention (SA) block by combining the latest depth prediction with the image feature to select useful image guidance features.
\item Extensive experiments on the benchmark datasets verify the superior effectiveness of the proposed framework and achieve state-of-the-art restoration performance.
\end{itemize}

\begin{figure*}[t]
  \centering
    \includegraphics[width=0.95\linewidth]{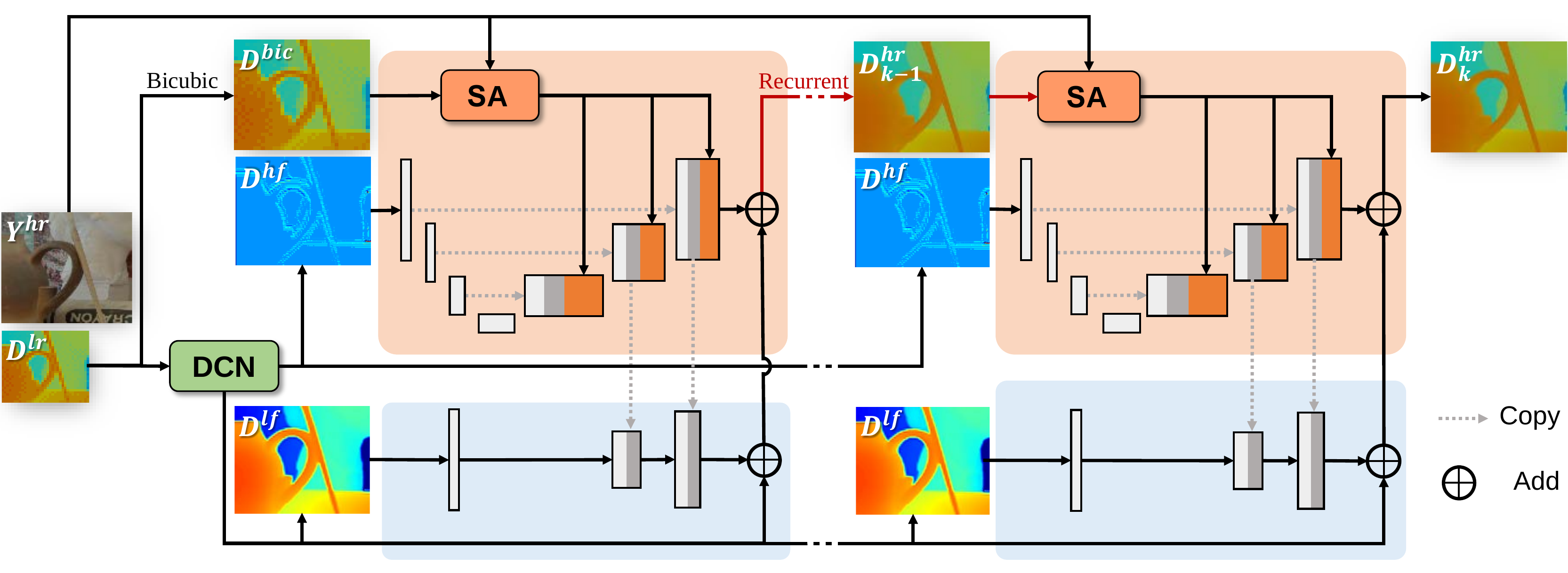}
    \caption{The pipeline of our RSAG framework. It consists of a green DCN module for the adaptive frequency-domain separation, an orange recurrent SA module for the HF component recovery, and a blue module for the LF component recovery. }
    \label{fig:network}
\end{figure*}

\section{Related work}
In this section, we mainly review the previous spectral decomposition and cross-modality fusion mechanisms used in depth map super-resolution (DSR). 

\subsection{Spectral Decomposition in DSR}
Since the HF component of the depth map can provide sufficient structure information which coincides well with the image boundaries, most methods adopt early spectral decomposition for efficient DSR. A line of methods \cite{makarov2017semi,xiao2018joint,li2019joint,zuo2019multi,2019Hierarchical} regard the interpolated depth input as the LF component and add a jump connection to transfer it to the end of the network. This global residual learning forces the network to focus on recovering the HF details. Another line of methods adopt the hand-designed filters \cite{2016Multi,yang2017depth} or edge-attention \cite{ye2018depthF,chen2018single} blocks to extract HF information. However, these methods require additional completion operation, since the HF outputs always include broken edges and holes. Recently, octave convolution \cite{chen2019drop} is utilized for frequency division operation in DSR network \cite{he2021towards}, which is a plug-and-play convolutional unit. However, it separates the frequency domain in embedding space, which does not guarantee that HF information is completely extracted. Instead, we propose a simple, fast, and adaptive separation method at the pixel level to provide reliable HF and LF maps.

\subsection{Cross-modality Fusion Mechanism}

\textbf{Multi-path/scale Learning.}
Previous methods \cite{li2016deep,lutio2019guided,zhu2018co,chen2018single,hao2019multi,su2019pixel} extract features in color space and depth space through two independent paths respectively, and transfer common structures through a joint branch. However, the multi-path methods may cause details missing since the cross-modality features are only fused in one specific layer. To handle the abovementioned problem, recent methods \cite{2016Multi,2019Hierarchical,he2021towards,zuo2019multi,zuo2019residual,yan2022learning} adopt a multi-scale fusion strategy to merge the cross-modality features at different levels. Although the multi-scale methods have achieved considerable performance, the coarse aggregation may cause texture copying and depth bleeding.

\textbf{Recursive Learning.}
In order to generate higher-level details without introducing excessive parameters, recursive learning repeatedly applies similar modules for progressive image reconstruction. Existing recursive DSR methods \cite{wen2019deep,yang2019depth,song2020channel} construct the depth map in a coarse-to-fine manner by regarding the previous crude depth output as the input of the DSR network. Even though the multi-supervision and residual learning avoid vanishing or exploding gradient problems to a certain extent, there still exists the risk of falling into a local optimum.
However, we propose a recurrent guidance for DSR, which considers the previous depth prediction as the guidance information for the next recursion. As the recursion progresses, the continuously refined guidance is a strong constraint for better choosing the image features. 
     
\textbf{Attention Mechanism.}
In recent years, the attention mechanism \cite{zhang2019self, guo2020closed, wang2021regularizing} has achieved significant improvements in the low-level vision field.
 In DSR task, Song \textit{et al.} \cite{song2020channel} utilize the channel attention to focus on HF depth. Meanwhile, Tang \textit{et al.} \cite{tang2021bridgenet} also design an HF attention bridge to extract the useful HF information during the depth estimation process and input it into the reconstruction network. 
Although these attention operations selectively highlight the HF information, they do not essentially solve the problem of texture copying and inconsistent boundaries in guidance images. The most related to our method is \cite{zhong2021high}, which also aims to find the consistent structure with an attention mechanism. However, there are big differences between them. $1)$ The proposed method uses contrastive networks to explore the cross-modality  correlation in the HF layer since the HF modalities of the depth map and image are closer.
$2)$ Compared with single image guidance, we complement the guidance with progressively refined depth prediction in a recursive fashion to accurately mine the consistent structure.

\section{Approach}
In this section, we introduce our recurrent structure attention guided (RSAG) framework for DSR. As shown in Fig.~\ref{fig:network}, RSAG contains three modules, including a deep contrastive network with multi-scale filters (DCN), a recurrent structure attention module (SA), and an HF\&LF feature fusion (HLF) module. DCN adaptively learns the HF and LF decomposition by cascading contrastive networks from large filters to small ones. 
Then, by introducing the last depth prediction to complement the image guidance, recurrent SA jointly selects the useful and clear structure features of the image for accurate HF depth reconstruction. Furthermore, during the reconstruction process, HF features guided by recurrent SA are integrated with LF features to refine the LF edges. 

Before presenting our method, we denote a high-resolution (HR) image by $Y^{hr}\in \mathbb{R}^{H\times W}$, where $H$ and $W$ are the sizes of the image, an HR depth map by $D^{hr}\in \mathbb{R}^{H\times W}$, a low-resolution (LR) depth map by $D^{lr}\in \mathbb{R}^{pH\times pW}$, where $0<p\leq 1$ is the downscaling factor (\textit{e.g.}, $1/4$, $1/8$, and $1/16$). For $D^{hr}$, $D^{lf}\in \mathbb{R}^{H\times W}$ and $D^{hf}\in \mathbb{R}^{H\times W}$ are denoted as its LF and HF components.

\subsection{Deep Contrastive Network with Multi-scale Filters}
 \label{sec:CFDS}
\begin{figure}[t]
  \centering
     \includegraphics[width=0.98\linewidth]{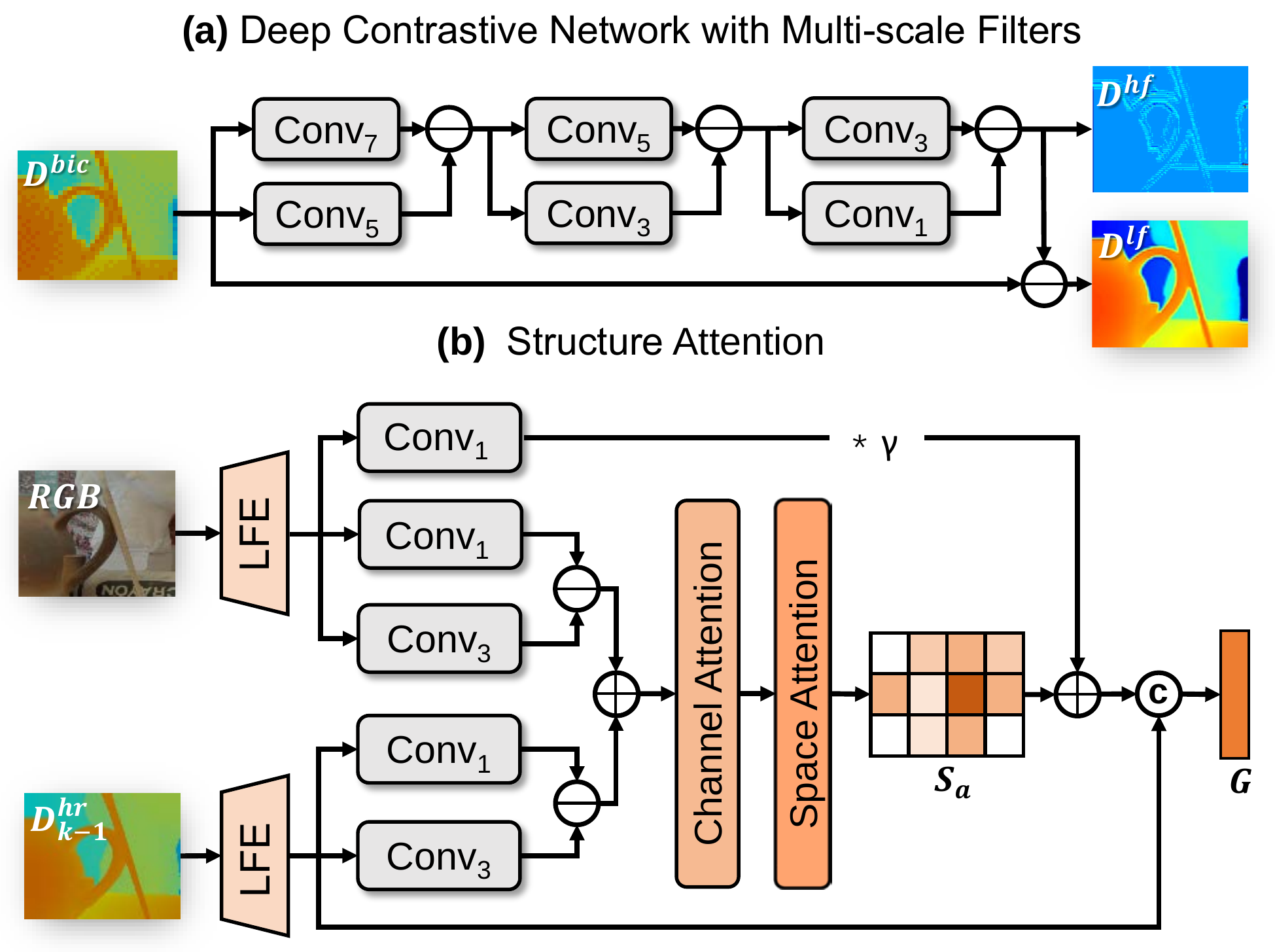}
   \caption{The architectures of (a) DCN and (b) SA. DCN aims to decompose HF and LF maps of a depth map by stacking three contrastive networks from large to small filters. SA tends to adaptively filter out unwanted textures and highlight the useful HF regions of the image. }
   \label{fig:subnetwork} 
\end{figure}

As shown in Fig.~\ref{fig:subnetwork} (a), we aim to explore a DCN network for high-quality frequency components, instead of the hand-designed operator for the frequency-domain decomposition. Inspired by the contrast learning operator \cite{xu2020learning}, which is designed for RGB image decomposition, we stack it three times to a DCN with multi-scale filters for extracting high-quality HF components of the depth map.

\begin{figure}[t]
 \centering
     \includegraphics[width=0.80\linewidth]{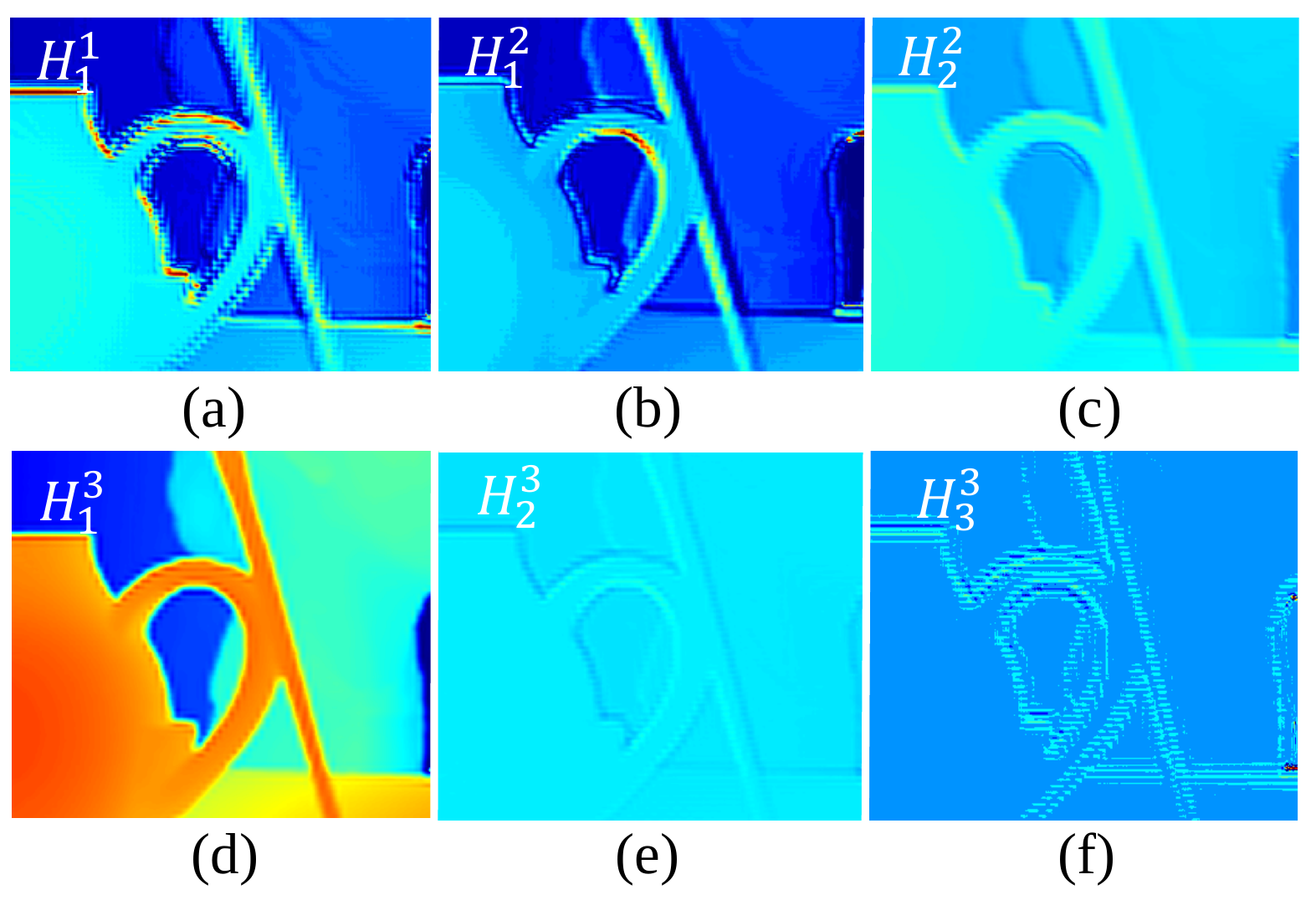}
   \caption{Visual HF features of our DCN network with different layers ($I=1,2,3$).
}
\label{fig:CFDSfeaturemap} 
\end{figure}

Specifically, given an LR depth map $D^{lr}\in \mathbb{R}^{pH\times pW}$ as input, we first upscale it to the desired resolution map $D^{bic}\in \mathbb{R}^{H\times W}$ by bicubic interpolation. We denote the number of layers of our DCN network as $I$, and the HF map $D^{hf}$ is defined as a recursive formulation:
\begin{equation}
D^{hf}=H_{I}^I, 
\end{equation}
\begin{equation}
 H_{i}^I=\text{Sigmoid}\left( \text{Conv}_{k} (H_{i-1}^I)-\text{Conv}_{k-2}(H_{i-1}^I) \right),
\end{equation}
where $H_0^I=D^{bic}$; $H_{i}^I$ is the HF feature of the $i$-th layer in the DCN network with $I$ layers ($1\leq i\leq I$); $\text{Conv}_{k}(\cdot)$ represents a $k\times k$ convolutional operation followed by PReLU~\cite{he2015delving} activation, $k =2(I-i)+3$, and we set $I=3$ in this paper. Then, the LF map $D^{lf}$ is calculated by subtracting the HF map $D^{hf}$ from $D^{bic}$:
\begin{equation}
   	 D^{lf}=  D^{bic}-D^{hf}.
\end{equation}
To better understand the DCN network with different layers ($I=1,2,3$), Fig.~\ref{fig:CFDSfeaturemap} shows their HF features. Fig.~\ref{fig:CFDSfeaturemap} (a-c) plot the HF features $H_{1}^1$, $H_{1}^2$ and $H_{2}^2$ of shallow DCN networks. Compared to $H_{1}^1$ and $H_{2}^2$, the HF feature $H_{3}^3$ of deeper DCN network are shown in Fig.~\ref{fig:CFDSfeaturemap} (f), which has the clearest and most complete edges.  According to Fig.~\ref{fig:CFDSfeaturemap} (d-f), it is worth noticing that deeper DCN is prone to weaken depth information and enhance structural information (e.g., edge of plaster statue behind the teapot).

 \subsection{Recurrent Structure Attention}
 \label{sec:CS}

Removing textures while making full use of consistent boundaries in the image is a key challenge for guided DSR. Instead of trivial cross-modality feature concatenation, we propose a novel recurrent structure attention (SA) mechanism to bridge the modality gap between depth input and image guidance.
As shown in Fig.~\ref{fig:subnetwork} (b), we put our efforts into the following two aspects:
(1) A cross-modality structure feature attention is designed, where the consistent structures are highlighted by contrast operators and the redundant features (e.g., textures and inconsistent boundaries) are suppressed in channel and space levels.
(2) For better guiding depth details restoration, useful image features are selected with the progressively refined depth prediction recursively. 

\textbf{Structure Attention.} 
Given the image $Y^{hr}\in \mathbb{R}^{H\times W}$ and the same size depth map $D^{hr}\in \mathbb{R}^{H\times W}$ as input, we first use the learnable feature extractor to produce a set of hierarchical image and depth features, which match with the corresponding  HF features in decoder path. Then, sharing the same spirit as contrastive networks used in DCN, we exploit the contextual information under multiple-level receptive fields and calculate the high contrastive features as HF components. We further sum the depth and image contrast maps and use absolute operations to enforce their consistent structures and prevent HF smoothing caused by positive and negative cancellations. This process can be formulated as:
\begin{align} 
J&=|F_{y_{1}}-F_{y_{2}}| + |F_{d_{1}}-F_{d_{2}}|, \\
F_{y_{i}}&=\text{Conv}_{2i-1}(\text{LFE}(Y^{hr})),\\
F_{d_{i}}&=\text{Conv}_{2i-1}(\text{LFE}(D^{hr})), i\in\left\{1,2\right\},
\end{align} 
where $\text{LFE}(\cdot)$ denotes the learnable feature extractor for initial hierarchical features learning. $\text{Conv}_{2i-1}(\cdot)$ are the convolutions with kernel size $2i-1$ followed by PReLU~\cite{he2015delving} activation. $F_{y_{i}}$ and $F_{d_{i}}$ are extracted image features and depth features under different receptive fields, respectively. $J$ denotes the joint HF features, which are further fed into the channel and spatial attention blocks \cite{woo2018cbam}. Such a design encourages learning the interaction between different channels and focusing on the important spatial locations. The features after the attention block denoted as structure-aware features $S_{a}$, can be formulated as:
\begin{equation}
S_{a}=\text{SpatA}(\text{CA}(J)),
\end{equation} 
where $\text{SpatA}(\cdot)$ and $\text{CA}(\cdot)$ represent the spatial attention and the channel attention blocks, respectively. At last, $S_{a}$ is added to the image features and combined with the depth features. The SA process is formulated as follows:
\begin{align}
G=&\text{SA}(D^{hr},Y^{hr}) \nonumber \\
=&\text{Cat}(\text{LFE}(D^{hr}), S_{a}+\gamma*\text{LFE}(\text{Conv}_{1}(Y^{hr}))),
\label{eq:fy}
\end{align} 
where $\gamma$ denotes a learnable parameter for controlling the degree of highlighting and $\text{Cat}(\cdot)$ means concatenation of features. $\text{Conv}_{1}(\cdot)$ is a $1\times1$ convolutional kernel followed by PReLU~\cite{he2015delving} activation. $G$ represents the fused guidance features for feeding into the decoder path.


\textbf{Recurrent Mechanism with Refined Depth Guidance.}
As mentioned above, compared to the single image guidance, the HR depth guidance owns the same modality as the LR depth input, which facilitates our attention module to locate and select consistent edge structures in image guidance. More clear depth structures can achieve more accurate guidance information for better details restoration, thence we refine the depth guidance in a recursive manner.
 
\begin{figure}
\centering
\includegraphics[width=0.80\linewidth]{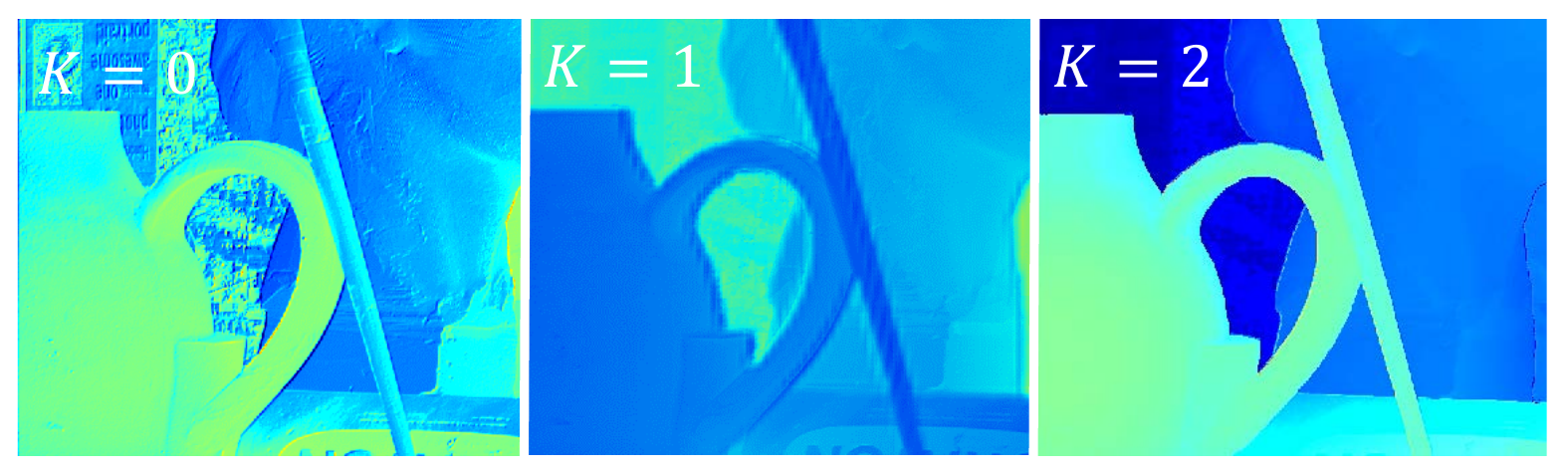}
\caption{Visual image features calculated by the Eq. \eqref{eq:fy} when the recursive step is varied from $k=0$ to $2$.}
\label{fig:featuremap} 
\end{figure}

Specifically, for the first recursion, the input depth guidance is the up-sampled version of LR depth map $D^{bic}\in \mathbb{R}^{H\times W}$ by bicubic interpolation, i.e., $G_{0}=\text{SA}(D^{bic},Y^{hr})$. For the $k$-th recursion, the latest output of our DSR network is taken as the input of the attention module next time. The recurrent SA can be formulated as follows:
\begin{align}
G_{k}=&\text{SA}(\text{HLF}(G_{k-1},D^{lf},D^{hf}),Y^{hr}),
\end{align} 
where $\text{HLF}(\cdot)$ is the HF\&LF feature fusion operation.
As shown in Fig.~\ref{fig:featuremap}, the image feature map before being inputted into the SA module contains complex patterns and unclear boundaries. As the recursion progresses, complex textures are removed (e.g., background pattern and cylinder label).

\subsection{HF\&LF feature fusion module}
 \label{sec:LF}  
Different from  previous methods directly up-sampling LF component by bicubic interpolation,
we propose an HF\&LF feature fusion (HLF) module to reconstruct the HF component and improve the blurred LF edges. 
The HF reconstruction module is built upon the  U-Net architecture, including an encoder path, an attention-based guidance branch, and a decoder branch (See orange blocks of  Fig.~\ref{fig:network}).
Rich hierarchical features extracted from the guidance branch and encoder-decoder structure are fused by using repeated residual convolutional block attention modules \cite{woo2018cbam,guo2020closed}. Then, the achieved contextual features in the decoder branch are concatenated with the LF features at multiple levels for the edges refining during the LF reconstruction (See blue blocks of  Fig.~\ref{fig:network}).

\subsection{Loss Function}
We train our model by minimizing the smooth-L1 loss between the network output $D^{hr}$ of each recursion and the ground-truth depth map $D^{gt}$. For the $k$-th recursion, the loss function $\mathcal{L}_{k}(\cdot)$ is defined as below:
\begin{align}
\ \mathcal{L}_{k}(D^{hr}_{k},D^{gt})= \sum_{i=1}^{N} \operatorname{smooth_{L1}}(D^{hr}_{k,i},D^{gt}_{i}),
\end{align}
where $ \operatorname{smooth_{L1}}(x) = \left\{\begin{matrix}
0.5x^{2}, &\text{if} \left | x \right |<1 \\
|x|-0.5,  &\text{otherwise}
\end{matrix}\right. $. $D^{hr}_{k}$ denotes the network output of the $k$-th recursion. $N$ and $i$ indicate the pixel number and the pixel index in the map, respectively. 
We can obtain $K$  depth outputs and the overall loss is expressed as:
	\begin{align}
	\ \mathcal{L}_{s} = \sum_{k=1}^{K}\lambda _{k}\mathcal{L}_{k},
	\end{align}
where $\lambda _{k}$ is the weight coefficient of the $k$-th loss. 

\section{Experiment}
\subsection{Experimental Setting}
To evaluate the performance of our framework, we conduct sufficient experiments on five datasets:
\begin{itemize}\setlength{\itemsep}{0pt} 
    \item Middlebury \cite{hirschmuller2007evaluation} \&  MPI Sintel \cite{butler2012naturalistic}: Training dataset consists of 34 RGB/D pairs from Middlebury dataset and 58 RGB/D pairs from MPI Sintel dataset. Testing dataset includes 6 RGB/D pairs (\textit{Art}, \textit{Books}, \textit{Dolls}, \textit{Laundry}, \textit{Mobeius}, \textit{Reindeer}) from Middlebury 2005.
    \item NYU-v2 \cite{silberman2012indoor}: Following the widely used data splitting manner, we sample 1000 pairs for training and the rest 449 pairs for testing. 
    \item Lu \cite{lu2014depth}: We test 6 RGB/D pairs from this dataset with the training model on NYU-v2.
    \item RGB-D-D \cite{he2021towards}: Following FDSR \cite{he2021towards}, we use 405 RGB/D pairs for evaluation  with the training model on NYU-v2. 
\end{itemize}

\begin{table*}[t]
	\centering
	\resizebox{0.95\linewidth}{!}{
\begin{tabular}{c|ccc|ccc|ccc|ccc|ccc|ccc}
\toprule[1.2pt]
 \multirow{2}{*}{Model} & \multicolumn{3}{c|}{\textit{Art}} &  \multicolumn{3}{c|}{\textit{Books}} & \multicolumn{3}{c|}{\textit{Dolls}} & \multicolumn{3}{c|}{\textit{Laundry}} & \multicolumn{3}{c|}{\textit{Mobeius}} & \multicolumn{3}{c}{\textit{Reindeer}}\\
\cline{2-19}
&$\times4$&$\times8$ &$\times16$
&$\times4$&$\times8$& $\times16$
&$\times4$&$\times8$& $\times16$
&$\times4$&$\times8$& $\times16$
&$\times4$&$\times8$& $\times16$
&$\times4$&$\times8$& $\times16$   \\ 
\midrule
Bicbuic &1.15  & 2.15 & 4.04 & 0.41 & 0.72 & 1.32 & 0.44 & 0.76 & 1.31 & 0.65 & 1.17 & 2.17 & 0.41 & 0.76 & 1.37 & 0.66 & 1.16 & 2.26 \\
DJF  &   0.40 &1.07& 2.78 &0.16& 0.45 &1.00& 0.20 &0.49& 0.99 &0.28& 0.71 &1.67& 0.18 &0.46& 1.02 &0.23& 0.60 &1.36\\
DMSG& 0.46 &0.76& 1.53 &0.15& 0.41 &0.76& 0.25 &0.51& 0.87 &0.30& 0.46 &1.12& 0.21 &0.43& 0.76 &0.31& 0.52 &0.99\\
DGDIE & 0.48 &1.20& 2.44 &0.30& 0.58 &1.02& 0.34 &0.63& 0.93 &0.35& 0.86 &1.56& 0.28 &0.58& 0.98 &0.35& 0.73& 1.29\\
GSPRT &  0.48 &0.74& 1.48 &0.21& 0.38 &0.76& 0.28 &0.48& 0.79 &0.33& 0.56 &1.24& 0.24 &0.49& 0.80 &0.31& 0.61 &1.07 \\
DJFR& 0.33& 0.71& 1.72 &0.19& 0.38 &0.78& 0.25 &0.44& 0.79 &0.22& 0.50 &1.12& 0.20 &0.38& 0.76 &0.24& 0.45 &0.96  \\
PacNet& 0.40 &0.82& 1.59 &0.22& 0.49 &0.84& 0.28 &0.53& 0.85 &0.28& 0.56 &1.08& 0.23 &0.44& 0.79 &0.29& 0.53 &1.00 \\
CUNet& 0.47 &1.06& 2.34 &0.33& 0.63 &1.41& 0.40 &0.67& 1.27 &0.41& 0.80 &1.88& 0.29 &0.65& 1.12 &0.35& 0.69 &1.14\\
PMBAN & 0.28 &0.55&1.11 &0.19& 0.30 &0.53& 0.23 &0.37& 0.64 &0.21& 0.36 &0.74& 0.18 &0.31& 0.57 &0.22& 0.39 &0.75\\
DKN&0.25 & 0.51& 1.22 &0.16& 0.30 &0.52& 0.21 &\underline{0.35}&  \underline{0.61} &0.17& \underline{0.34} &0.81& 0.16 &0.28& 0.54 &0.20& 0.38 &0.70\\
AHMF& \underline{0.22}& \underline{0.50} & \underline{1.04}& 0.14& 0.30&  \textbf{0.50}& \underline{0.18}& \underline{0.35} &0.62& \underline{0.15}&  \underline{0.34}& \underline{0.73}& 0.14& 0.28& \underline{0.53}& 0.18& 0.37 &\underline{0.64}\\
CTKT&0.25&0.53&1.44&\underline{0.11}&\underline{0.26}&0.67&0.16&0.36&0.65&0.16&0.36&0.76&\underline{0.13}&\underline{0.27}&0.69&\underline{0.17}&\underline{0.35}&0.77\\
\midrule
RSAG& \textbf{0.13}&\textbf{0.23}&\textbf{0.88}&\textbf{0.09}&\textbf{0.14}&\textbf{0.50}& \textbf{0.15}&\textbf{0.20}&\textbf{0.57}&\textbf{0.10}&\textbf{0.19}&\textbf{0.58}&\textbf{0.12}&\textbf{0.17}&\textbf{0.42}&\textbf{0.13}&\textbf{0.18}&\textbf{0.52}\\
\bottomrule[1.2pt]
\end{tabular}}
\caption{Quantitative comparisons (in MAD) on Middlebury dataset.}
\label{tab:Middlebury}
\end{table*}

\begin{table*}[t]
	\centering
	\resizebox{0.95\linewidth}{!}{
\begin{tabular}{c|cccccccccccccc}
\toprule[1.2pt] 
&Bicubic &TGV&DJF &FBS& DMSG &DJFR  & GbFT &PacNet & FDKN  & DKN& FDSR &CTKT&DCTNet & RSAG\\
\midrule
$\times4$&8.16& 4.98 &3.54&4.29& 3.02   &2.38  &3.35  & 2.39 & 1.86 & 1.62 & 1.61 &\underline{1.49}  &1.59& \textbf{1.23}\\
$\times8$&14.22& 11.23 &6.20 &8.94&2.99  &4.94  &5.73  & 4.59 & 3.58  & 3.26 & 3.18 & \underline{2.73}   &3.16& \textbf{2.51}\\
$\times16$&22.32&28.13&10.21 &14.59& 9.17 &9.18 &9.01 & 8.09 & 6.96 & 6.51 & 5.86 &\textbf{5.11}   &5.84& \underline{5.27}\\
\bottomrule[1.2pt]
\end{tabular}}
\caption{Quantitative comparisons (in  RMSE (cm)) on NYU-v2 dataset.}\label{tab:NYU-v2}
\end{table*}

\begin{figure*}[t]
\centering
\includegraphics[width=0.98\linewidth]{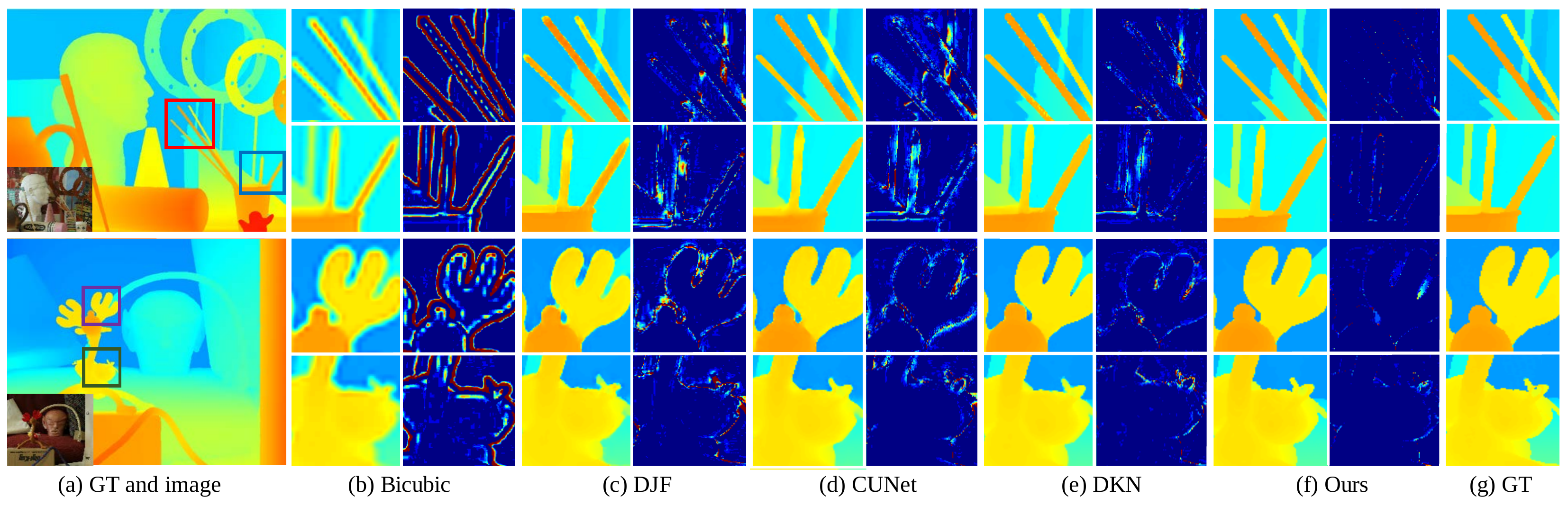}
\caption{Visual comparisons of \textit{Art} and \textit{Laundry} on  Middlebury dataset ($\times8$ case).} 
\label{fig:Middlebury}
\end{figure*}

We compare our method with 3 traditional methods: TGV \cite{ferstl2013image}, FBS \cite{barron2016fast}, SDF \cite{ham2017robust}, 3 classical methods: DJF  \cite{li2016deep}, DMSG \cite{2016Multi}, DGDIE \cite{gu2017learning} and 12 state-of-the-art (SOTA) methods: SVLRM \cite{pan2019spatially}, GSPRT \cite{lutio2019guided}, DJFR \cite{li2019joint}, PacNet \cite{su2019pixel}, GbFT  \cite{albahar2019guided}, CUNet \cite{deng2020deep}, PMBAN \cite{ye2020pmbanet}, DKN \cite{kim2021deformable}, FDKN \cite{kim2021deformable},  FDSR \cite{he2021towards}, AHMF \cite{zhong2021high} and   CTKT \cite{sun2021learning}. Mean Absolute Error (MAD) and Root Mean Squared Error (RMSE) are used to evaluate the performance.

During training, we randomly extract patches with stride = $\{96,96,128\}$ for the scale = $\{4,8,16\}$ respectively as ground truth and use bicubic interpolation to get LR inputs. The training and testing data are normalized to the range $[0,1]$. 
To balance the training time and network performance, we set the recurrent steps of the SA blocks as $k=2$ in this paper.
The loss weights are set as $\lambda _{k}=0.5$. The proposed method is implemented using PyTorch with one RTX 2080Ti GPU. For simplicity, we name our \textbf{R}ecurrent \textbf{S}tructure \textbf{A}ttention \textbf{G}uided framework as \textbf{RSAG}.

\subsection{Comparing to State-of-the-Arts}
\subsubsection{Quantitative Comparisons.} 
We first show the quantitative evaluation results with SOTA methods under the same conditions. Table~\ref{tab:Middlebury} shows the results on Middlebury dataset under three up-scaling factors. It can be observed that the proposed RSAG outperforms the SOTA methods by significant margins for all up-scaling factors. For example, RSAG decreases the average MAD by $25\% (\times4)$, $48\% (\times8)$, and $30\% (\times16)$ compared to  CTKT \cite{sun2021learning}. We further evaluate the proposed method on NYU-v2 dataset in Table~\ref{tab:NYU-v2}. The proposed method yields the best performance for $\times4$ and $\times8$ DSR and comparable performance for $\times16$ DSR. Compared with the second-best method, RSAG decreases the average RMSE by $17\%$ for $\times4$ DSR.

\begin{table}[t]
	\centering
	\resizebox{0.80\linewidth}{!}{
\begin{tabular}{c|ccc|ccc}
\toprule[1.2pt]
 \multirow{2}{*}{Model} & \multicolumn{3}{c|}{Lu } &  \multicolumn{3}{c}{RGB-D-D}\\
 \cline{2-7}
&$\times4$&$\times8$ &$\times16$&$\times4$&$\times8$ &$\times16$\\
\midrule
 DJF&1.65&3.96&6.75&3.41&5.57&8.15\\
 DJFR&1.15&3.57&6.77&3.35&5.57&7.99\\
 FDKN&0.82&2.10&5.05&1.18&1.91&3.41\\
 DKN &0.96 &2.16&5.11 &1.30&1.96&3.42 \\
 FDSR&\underline{0.81} &\underline{1.91}&\underline{4.64}&\underline{1.16} &\underline{1.82}&\underline{3.06}\\
 \midrule
 RSAG&\textbf{0.79}&\textbf{1.67}&\textbf{4.30}&\textbf{1.14}&\textbf{1.75}&\textbf{2.96}\\
\bottomrule[1.2pt]
\end{tabular}}
\caption{Quantitative comparisons (in RMSE) on Lu dataset and RGB-D-D dataset.}\label{tab:RGB-D-D}
\end{table}

To verify the generalization ability of our method on Lu dataset and RGB-D-D dataset, we test RSAG for $\times4$, $\times8$, and $\times16$ DSR, which is trained on NYU dataset. As shown in Table~\ref{tab:RGB-D-D}, we can see that RSAG performs the competitive generalization results for all up-sampling cases, which demonstrates the accuracy and effectiveness of our method.

\subsubsection{Visual Comparisons.} 
\begin{figure}[ht]
  \centering
    \includegraphics[width=0.98\linewidth]{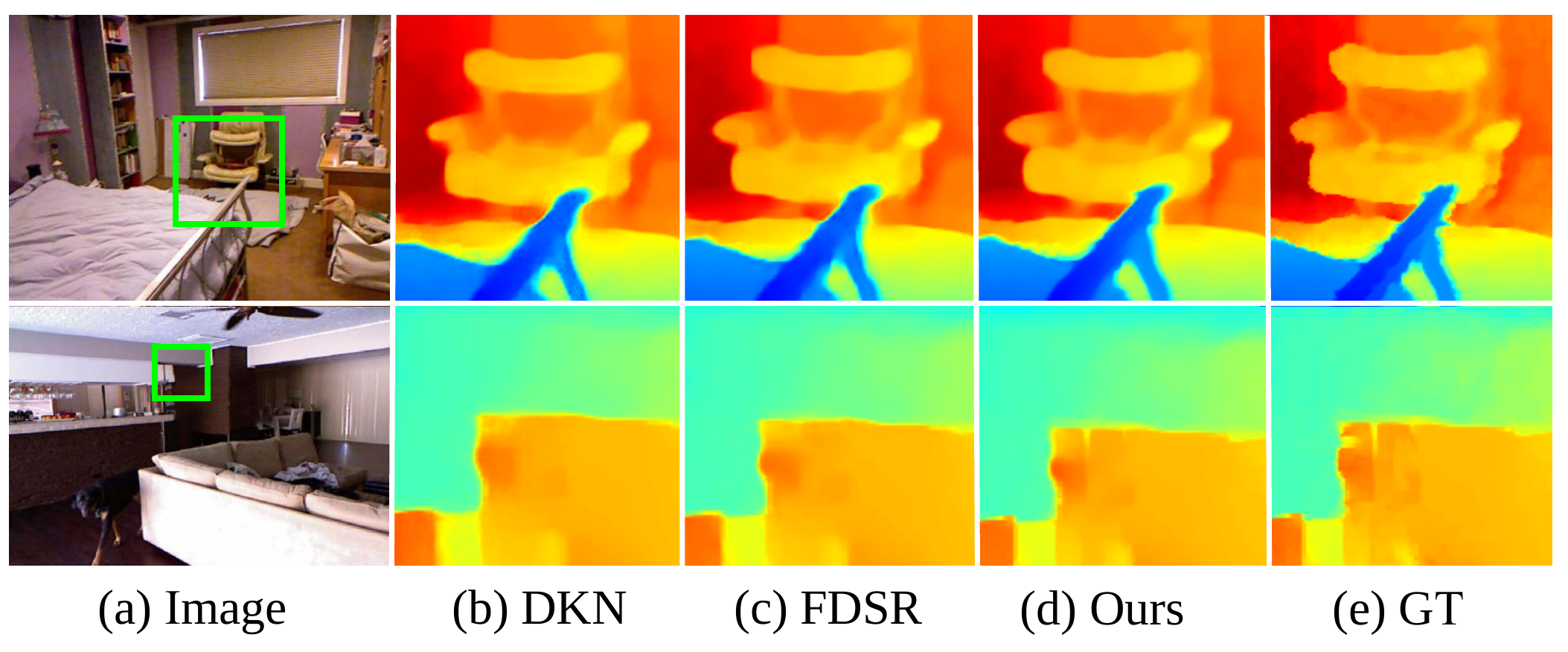}
    \caption{Visual comparisons on NYU-v2 dataset ($\times8$ case).} 
    \label{fig:resultNYU-v2}
\end{figure}

We provide the visual comparisons of the $\times8$ upsampled results on Middlebury dataset in Fig.~\ref{fig:Middlebury}.
It is worth noted that edges and luxuriant details are hard to be reconstructed by interpolation or simple feature concatenation. Even though CUNet \cite{deng2020deep} and DKN \cite{kim2021deformable} can recover most boundaries, they fail to reconstruct some complex structures, such as texture beside pencils in \textit{Art} and boundaries of antlers in \textit{Reindeer}. In contrast, our results show sharper edges and smaller errors with the ground truth.
Fig.~\ref{fig:resultNYU-v2} shows $\times8$ results on NYU-v2 dataset. Boundaries and details generated by RSAG are more accurate without introducing the texture copying artifacts, which demonstrates that RSAG can well recover both HF structures and LF content.

\begin{figure}[t]
  \centering
     \includegraphics[width=0.98\linewidth]{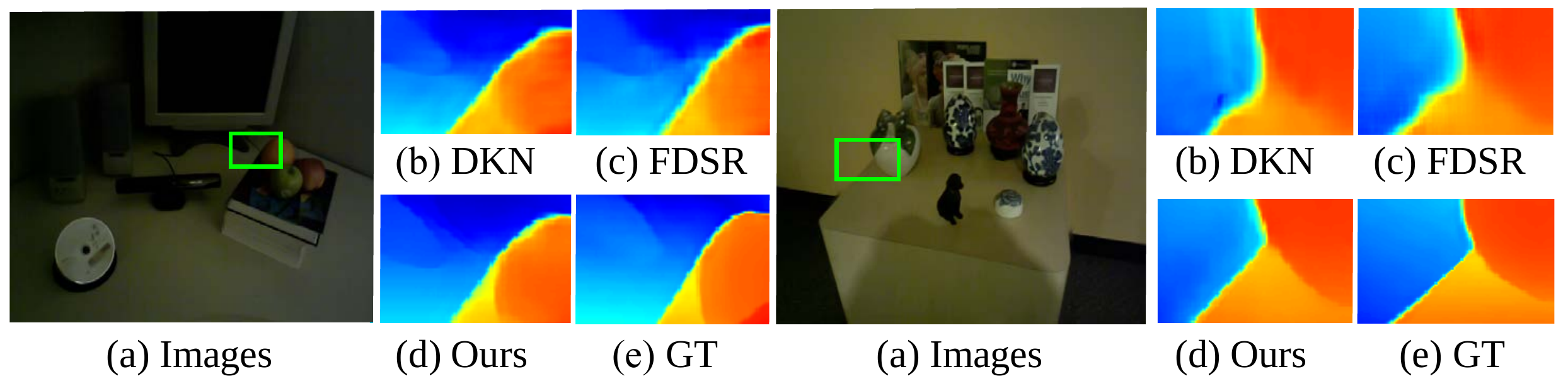}
   \caption{Visual comparisons on Lu dataset ($\times16$ case).} 
    \label{fig:Lu}
\end{figure}

Furthermore, Fig.~\ref{fig:Lu} demonstrates the good generalization ability of the proposed method on Lu dataset for $\times16$ DSR. Most methods generally tend to over-smooth the results and fail to recover the depth details with low-light guidance images, while our method produces more convincing results.

\begin{table}[t]
	\centering
	\resizebox{0.90\linewidth}{!}{
\begin{tabular}{lcc}
\toprule[1.2pt]
Model&\multicolumn{1}{|c}\textit{Middlebury} &{NYU-v2}\\
\midrule
baseline&
\multicolumn{1}{|c}{0.26}&3.60\\
baseline + DCN&
\multicolumn{1}{|c}{0.24}&3.10\\
baseline + DCN + HLF&
\multicolumn{1}{|c}{0.23}&3.02\\
baseline + DCN + HLF + SA&
\multicolumn{1}{|c}{0.19}&2.51\\
\bottomrule[1.2pt]
\end{tabular}}
\caption{Ablation studies of RSAG (in MAD) on Middlebury dataset and (in RMSE) on  NYU-v2 dataset  for $\times8$ DSR.}
\label{tab:Ablation1}
\end{table}

\subsection{Ablation Study}
\label{Ablation}
\textbf{Effect of DCN and HLF modules.}
Table~\ref{tab:Ablation1} reports the ablation studies on the DCN and HLF modules in our framework. As shown in the first row of Table~\ref{tab:Ablation1}, the baseline model uses a hand-designed operator for frequency-domain decomposition and direct concatenation for cross-modality feature fusion. The second row demonstrates that the proposed DCN module, which selects HF component adaptively in a coarse-to-fine manner, can significantly improve the performance over the baseline. When the HLF module is added, the average RMSE of the NYU-v2 dataset shown in the third row can be reduced from 3.60 to 3.02, which further verifies the effectiveness of high-quality frequency-domain separation and HF\&LF feature fusion modules.

\begin{figure}[t]
  \centering
     \includegraphics[width=0.98\linewidth]{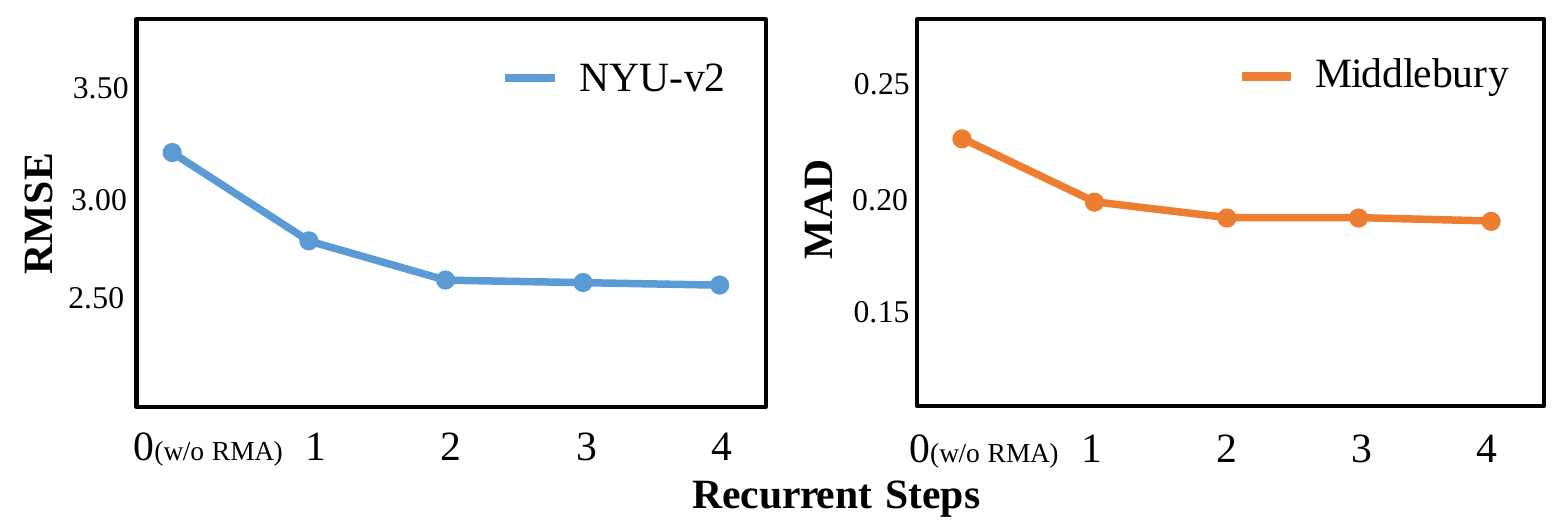}
   \caption{
  Ablation studies of SA with different recursive steps on Middlebury and NYU-v2 datasets ($\times8$ case).}
    \label{fig:number}
\end{figure}

\textbf{Effect of SA module.}
The last row in Table~\ref{tab:Ablation1} demonstrates the effectiveness of the SA module, which iteratively utilizes the latest depth estimation to choose clear and consistent image features. We can see that the SA module can outperform them by a large margin. 
From the results of Table~\ref{tab:Ablation1}, it is observed that all the modules proposed in the RSAG framework have made a positive contribution to the ultimate success of our method. To further study the impact of the recurrent steps of SA, we conduct experiments on Middlebury and NYU-v2 datasets by varying the step from 0 (w/o SA) to 4, as illustrated in Fig.~\ref{fig:number}.
It can be found that the method achieves better performance when the recursion steps increase, where 2 recurrent steps obtain the best trade-off between speed and accuracy.
It also proves that higher-quality depth information can help obtain a more reliable guidance structure for subsequent depth reconstruction.

\section{Conclusion}
\label{section:6}
In this paper, we proposed a novel recurrent structure attention guided (RSAG) framework for depth super-resolution. In our framework, a deep contrastive network with multi-scale filters (DCN) block was designed to adaptively decompose the high-quality HF and LF components by using contrastive networks from large kernels to small ones. In addition, by leveraging the latest depth output and high-resolution image as guidance, we introduced recurrent structure attention (SA) block, instead of the trivial feature concatenation, to select consistent and clear image features for subsequent cross-modality fusion.
Furthermore, we presented the HF\&LF feature fusion block to refine the blurred edges of the LF component. Extensive experiments on various benchmark datasets demonstrated the superiority and effectiveness of the proposed framework.

\section{Acknowledgement}
This work was supported by the National Science Fund of China under Grant Nos.~U1713208 and 62072242.

\bibliography{aaai23}

\begin{thebibliography}{46}
\providecommand{\natexlab}[1]{#1}

\bibitem[{AlBahar and Huang(2019)}]{albahar2019guided}
AlBahar, B.; and Huang, J.-B. 2019.
\newblock Guided image-to-image translation with bi-directional feature
  transformation.
\newblock In \emph{ICCV}, 9016--9025.

\bibitem[{Barron and Poole(2016)}]{barron2016fast}
Barron, J.~T.; and Poole, B. 2016.
\newblock The fast bilateral solver.
\newblock In \emph{ECCV}, 617--632.

\bibitem[{Butler et~al.(2012)Butler, Wulff, Stanley, and
  Black}]{butler2012naturalistic}
Butler, D.~J.; Wulff, J.; Stanley, G.~B.; and Black, M.~J. 2012.
\newblock A naturalistic open source movie for optical flow evaluation.
\newblock In \emph{ECCV}, 611--625.

\bibitem[{Caesar et~al.(2020)Caesar, Bankiti, Lang, Vora, Liong, Xu, Krishnan,
  Pan, Baldan, and Beijbom}]{caesar2020nuscenes}
Caesar, H.; Bankiti, V.; Lang, A.~H.; Vora, S.; Liong, V.~E.; Xu, Q.; Krishnan,
  A.; Pan, Y.; Baldan, G.; and Beijbom, O. 2020.
\newblock nuscenes: A multimodal dataset for autonomous driving.
\newblock In \emph{CVPR}, 11621--11631.

\bibitem[{Chen and Jung(2018)}]{chen2018single}
Chen, B.; and Jung, C. 2018.
\newblock Single depth image super-resolution using convolutional neural
  networks.
\newblock In \emph{ICASSP}, 1473--1477.

\bibitem[{Chen et~al.(2019)Chen, Fan, Xu, Yan, Kalantidis, Rohrbach, Yan, and
  Feng}]{chen2019drop}
Chen, Y.; Fan, H.; Xu, B.; Yan, Z.; Kalantidis, Y.; Rohrbach, M.; Yan, S.; and
  Feng, J. 2019.
\newblock Drop an octave: Reducing spatial redundancy in convolutional neural
  networks with octave convolution.
\newblock In \emph{ICCV}, 3435--3444.

\bibitem[{Deng and Dragotti(2020)}]{deng2020deep}
Deng, X.; and Dragotti, P.~L. 2020.
\newblock Deep convolutional neural network for multi-modal image restoration
  and fusion.
\newblock \emph{IEEE transactions on pattern analysis and machine
  intelligence}, PP(99): 1--1.

\bibitem[{Ferstl et~al.(2013)Ferstl, Reinbacher, Ranftl, R{\"u}ther, and
  Bischof}]{ferstl2013image}
Ferstl, D.; Reinbacher, C.; Ranftl, R.; R{\"u}ther, M.; and Bischof, H. 2013.
\newblock Image guided depth upsampling using anisotropic total generalized
  variation.
\newblock In \emph{ICCV}, 993--1000.

\bibitem[{Gu et~al.(2017)Gu, Zuo, Guo, Chen, Chen, and Zhang}]{gu2017learning}
Gu, S.; Zuo, W.; Guo, S.; Chen, Y.; Chen, C.; and Zhang, L. 2017.
\newblock Learning dynamic guidance for depth image enhancement.
\newblock In \emph{CVPR}, 3769--3778.

\bibitem[{Guo et~al.(2019)Guo, Li, Guo, Cong, Fu, and Han}]{2019Hierarchical}
Guo, C.; Li, C.; Guo, J.; Cong, R.; Fu, H.; and Han, P. 2019.
\newblock Hierarchical Features Driven Residual Learning for Depth Map
  Super-Resolution.
\newblock \emph{IEEE Transactions on Image Processing}, 2545--2557.

\bibitem[{Guo et~al.(2020)Guo, Chen, Wang, Chen, Cao, Deng, Xu, and
  Tan}]{guo2020closed}
Guo, Y.; Chen, J.; Wang, J.; Chen, Q.; Cao, J.; Deng, Z.; Xu, Y.; and Tan, M.
  2020.
\newblock Closed-loop matters: Dual regression networks for single image
  super-resolution.
\newblock In \emph{CVPR}, 5407--5416.

\bibitem[{Ham, Cho, and Ponce(2017)}]{ham2017robust}
Ham, B.; Cho, M.; and Ponce, J. 2017.
\newblock Robust guided image filtering using nonconvex potentials.
\newblock \emph{IEEE transactions on pattern analysis and machine
  intelligence}, 40(1): 192--207.

\bibitem[{Hao et~al.(2019)Hao, Lu, Zhang, Wang, and Chen}]{hao2019multi}
Hao, X.; Lu, T.; Zhang, Y.; Wang, Z.; and Chen, H. 2019.
\newblock Multi-Source Deep Residual Fusion Network for Depth Image
  Super-resolution.
\newblock In \emph{RSVT}, 62--67.

\bibitem[{He et~al.(2015)He, Zhang, Ren, and Sun}]{he2015delving}
He, K.; Zhang, X.; Ren, S.; and Sun, J. 2015.
\newblock Delving deep into rectifiers: Surpassing human-level performance on
  imagenet classification.
\newblock In \emph{ICCV}, 1026--1034.

\bibitem[{He et~al.(2021)He, Zhu, Li, Bai, Cong, Zhang, Lin, Liu, and
  Zhao}]{he2021towards}
He, L.; Zhu, H.; Li, F.; Bai, H.; Cong, R.; Zhang, C.; Lin, C.; Liu, M.; and
  Zhao, Y. 2021.
\newblock Towards Fast and Accurate Real-World Depth Super-Resolution:
  Benchmark Dataset Baseline and.
\newblock In \emph{CVPR}, 9229--9238.

\bibitem[{Hirschmuller and Scharstein(2007)}]{hirschmuller2007evaluation}
Hirschmuller, H.; and Scharstein, D. 2007.
\newblock Evaluation of cost functions for stereo matching.
\newblock In \emph{CVPR}, 1--8.

\bibitem[{Hou, Dai, and Nie{\ss}ner(2019)}]{hou20193d}
Hou, J.; Dai, A.; and Nie{\ss}ner, M. 2019.
\newblock 3d-sis: 3d semantic instance segmentation of rgb-d scans.
\newblock In \emph{CVPR}, 4421--4430.

\bibitem[{Hui, Loy, and Tang(2016)}]{2016Multi}
Hui, T.-W.; Loy, C.~C.; and Tang, X. 2016.
\newblock Depth Map Super-Resolution by Deep Multi-Scale Guidance.
\newblock In \emph{ECCV}, 353--369.

\bibitem[{Kim, Ponce, and Ham(2021)}]{kim2021deformable}
Kim, B.; Ponce, J.; and Ham, B. 2021.
\newblock Deformable kernel networks for joint image filtering.
\newblock \emph{International Journal of Computer Vision}, 129(2): 579--600.

\bibitem[{Li et~al.(2016)Li, Huang, Ahuja, and Yang}]{li2016deep}
Li, Y.; Huang, J.-B.; Ahuja, N.; and Yang, M.-H. 2016.
\newblock Deep joint image filtering.
\newblock In \emph{ECCV}, 154--169.

\bibitem[{Li et~al.(2019)Li, Huang, Ahuja, and Yang}]{li2019joint}
Li, Y.; Huang, J.-B.; Ahuja, N.; and Yang, M.-H. 2019.
\newblock Joint image filtering with deep convolutional networks.
\newblock \emph{IEEE transactions on pattern analysis and machine
  intelligence}, 41(8): 1909--1923.

\bibitem[{Lu, Ren, and Liu(2014)}]{lu2014depth}
Lu, S.; Ren, X.; and Liu, F. 2014.
\newblock Depth enhancement via low-rank matrix completion.
\newblock In \emph{CVPR}, 3390--3397.

\bibitem[{Lutio et~al.(2019)Lutio, D'aronco, Wegner, and
  Schindler}]{lutio2019guided}
Lutio, R.~d.; D'aronco, S.; Wegner, J.~D.; and Schindler, K. 2019.
\newblock Guided super-resolution as pixel-to-pixel transformation.
\newblock In \emph{ICCV}, 8829--8837.

\bibitem[{Makarov, Aliev, and Gerasimova(2017)}]{makarov2017semi}
Makarov, I.; Aliev, V.; and Gerasimova, O. 2017.
\newblock Semi-dense depth interpolation using deep convolutional neural
  networks.
\newblock In \emph{ACMMM}, 1407--1415.

\bibitem[{Meuleman et~al.(2020)Meuleman, Baek, Heide, and
  Kim}]{meuleman2020single}
Meuleman, A.; Baek, S.-H.; Heide, F.; and Kim, M.~H. 2020.
\newblock Single-shot monocular rgb-d imaging using uneven double refraction.
\newblock In \emph{CVPR}, 2465--2474.

\bibitem[{Pan et~al.(2019)Pan, Dong, Ren, Lin, Tang, and
  Yang}]{pan2019spatially}
Pan, J.; Dong, J.; Ren, J.~S.; Lin, L.; Tang, J.; and Yang, M.-H. 2019.
\newblock Spatially variant linear representation models for joint filtering.
\newblock In \emph{CVPR}, 1702--1711.

\bibitem[{Silberman et~al.(2012)Silberman, Hoiem, Kohli, and
  Fergus}]{silberman2012indoor}
Silberman, N.; Hoiem, D.; Kohli, P.; and Fergus, R. 2012.
\newblock Indoor segmentation and support inference from rgbd images.
\newblock In \emph{ECCV}, 746--760.

\bibitem[{Song et~al.(2020)Song, Dai, Zhou, Liu, Li, Li, and
  Yang}]{song2020channel}
Song, X.; Dai, Y.; Zhou, D.; Liu, L.; Li, W.; Li, H.; and Yang, R. 2020.
\newblock Channel attention based iterative residual learning for depth map
  super-resolution.
\newblock In \emph{CVPR}, 5631--5640.

\bibitem[{Su et~al.(2019)Su, Jampani, Sun, Gallo, Learned-Miller, and
  Kautz}]{su2019pixel}
Su, H.; Jampani, V.; Sun, D.; Gallo, O.; Learned-Miller, E.; and Kautz, J.
  2019.
\newblock Pixel-adaptive convolutional neural networks.
\newblock In \emph{CVPR}, 11166--11175.

\bibitem[{Sun et~al.(2021)Sun, Ye, Li, Li, Wang, and Xu}]{sun2021learning}
Sun, B.; Ye, X.; Li, B.; Li, H.; Wang, Z.; and Xu, R. 2021.
\newblock Learning Scene Structure Guidance via Cross-Task Knowledge Transfer
  for Single Depth Super-Resolution.
\newblock In \emph{CVPR}, 7792--7801.

\bibitem[{Tang et~al.(2021)Tang, Cong, Sheng, He, Zhang, Zhao, and
  Kwong}]{tang2021bridgenet}
Tang, Q.; Cong, R.; Sheng, R.; He, L.; Zhang, D.; Zhao, Y.; and Kwong, S. 2021.
\newblock BridgeNet: A Joint Learning Network of Depth Map Super-Resolution and
  Monocular Depth Estimation.
\newblock In \emph{ACMMM}, 2148--2157.

\bibitem[{Wang et~al.(2021)Wang, Zhang, Yan, Li, Xu, Li, and
  Yang}]{wang2021regularizing}
Wang, K.; Zhang, Z.; Yan, Z.; Li, X.; Xu, B.; Li, J.; and Yang, J. 2021.
\newblock Regularizing nighttime weirdness: Efficient self-supervised monocular
  depth estimation in the dark.
\newblock In \emph{ICCV}, 16055--16064.

\bibitem[{Wen et~al.(2019)Wen, Sheng, Li, Lin, and Feng}]{wen2019deep}
Wen, Y.; Sheng, B.; Li, P.; Lin, W.; and Feng, D.~D. 2019.
\newblock Deep color guided coarse-to-fine convolutional network cascade for
  depth image super-resolution.
\newblock \emph{IEEE Transactions on Image Processing}, 28(2): 994--1006.

\bibitem[{Woo et~al.(2018)Woo, Park, Lee, and Kweon}]{woo2018cbam}
Woo, S.; Park, J.; Lee, J.-Y.; and Kweon, I.~S. 2018.
\newblock Cbam: {C}onvolutional block attention module.
\newblock In \emph{ECCV}, 3--19.

\bibitem[{Xiao et~al.(2018)Xiao, Cao, Zhu, Yang, and Zheng}]{xiao2018joint}
Xiao, Y.; Cao, X.; Zhu, X.; Yang, R.; and Zheng, Y. 2018.
\newblock Joint convolutional neural pyramid for depth map super-resolution.
\newblock \emph{arXiv preprint arXiv:1801.00968}.

\bibitem[{Xu et~al.(2020)Xu, Yang, Yin, and Lau}]{xu2020learning}
Xu, K.; Yang, X.; Yin, B.; and Lau, R.~W. 2020.
\newblock Learning to restore low-light images via
  decomposition-and-enhancement.
\newblock In \emph{CVPR}, 2281--2290.

\bibitem[{Yan et~al.(2022)Yan, Wang, Li, Zhang, Li, Li, and
  Yang}]{yan2022learning}
Yan, Z.; Wang, K.; Li, X.; Zhang, Z.; Li, G.; Li, J.; and Yang, J. 2022.
\newblock Learning Complementary Correlations for Depth Super-Resolution With
  Incomplete Data in Real World.
\newblock \emph{IEEE Transactions on Neural Networks and Learning Systems}.

\bibitem[{Yang et~al.(2019)Yang, Fan, Zheng, Liu, Zhang, and
  Lei}]{yang2019depth}
Yang, B.; Fan, X.; Zheng, Z.; Liu, X.; Zhang, K.; and Lei, J. 2019.
\newblock Depth Map Super-Resolution via Multilevel Recursive Guidance and
  Progressive Supervision.
\newblock \emph{IEEE Access}, 7: 57616--57622.

\bibitem[{Yang et~al.(2017)Yang, Lan, Song, and Li}]{yang2017depth}
Yang, J.; Lan, H.; Song, X.; and Li, K. 2017.
\newblock Depth super-resolution via fully edge-augmented guidance.
\newblock In \emph{VCIP}, 1--4.

\bibitem[{Ye, Duan, and Li(2018)}]{ye2018depthF}
Ye, X.; Duan, X.; and Li, H. 2018.
\newblock Depth super-resolution with deep edge-inference network and
  edge-guided depth filling.
\newblock In \emph{ICASSP}, 1398--1402.

\bibitem[{Ye et~al.(2020)Ye, Sun, Wang, Yang, Xu, Li, and Li}]{ye2020pmbanet}
Ye, X.; Sun, B.; Wang, Z.; Yang, J.; Xu, R.; Li, H.; and Li, B. 2020.
\newblock Pmbanet: {P}rogressive multi-branch aggregation network for scene
  depth super-resolution.
\newblock \emph{IEEE Transactions on Image Processing}, 29: 7427--7442.

\bibitem[{Zhang et~al.(2019)Zhang, Goodfellow, Metaxas, and
  Odena}]{zhang2019self}
Zhang, H.; Goodfellow, I.; Metaxas, D.; and Odena, A. 2019.
\newblock Self-attention generative adversarial networks.
\newblock In \emph{ICML}, 7354--7363.

\bibitem[{Zhong et~al.(2021)Zhong, Liu, Jiang, Zhao, Chen, and
  Ji}]{zhong2021high}
Zhong, Z.; Liu, X.; Jiang, J.; Zhao, D.; Chen, Z.; and Ji, X. 2021.
\newblock High-resolution Depth Maps Imaging via Attention-based Hierarchical
  Multi-modal Fusion.
\newblock \emph{arXiv preprint arXiv:2104.01530}.

\bibitem[{Zhu et~al.(2018)Zhu, Zhai, Cao, and Zha}]{zhu2018co}
Zhu, J.; Zhai, W.; Cao, Y.; and Zha, Z.-J. 2018.
\newblock Co-occurrent structural edge detection for color-guided depth map
  super-resolution.
\newblock In \emph{MMM}, 93--105.

\bibitem[{Zuo et~al.(2019{\natexlab{a}})Zuo, Fang, Yang, Shang, and
  Wang}]{zuo2019residual}
Zuo, Y.; Fang, Y.; Yang, Y.; Shang, X.; and Wang, B. 2019{\natexlab{a}}.
\newblock Residual dense network for intensity-guided depth map enhancement.
\newblock \emph{Information Sciences}, 495: 52--64.

\bibitem[{Zuo et~al.(2019{\natexlab{b}})Zuo, Wu, Fang, An, Huang, and
  Chen}]{zuo2019multi}
Zuo, Y.; Wu, Q.; Fang, Y.; An, P.; Huang, L.; and Chen, Z. 2019{\natexlab{b}}.
\newblock Multi-scale frequency reconstruction for guided depth map
  super-resolution via deep residual network.
\newblock \emph{IEEE Transactions on Circuits and Systems for Video
  Technology}, 30(2): 297--306.

\end{thebibliography}

\end{document}